\documentclass[final]{cvpr}

\usepackage{times}
\usepackage{epsfig}
\usepackage{graphicx}
\usepackage{amsmath}
\usepackage{amssymb}
\usepackage{subfiles}
\usepackage{multirow} 
\newcommand{\ra}[1]{\renewcommand{\arraystretch}{#1}}
\usepackage{booktabs}
\usepackage[ruled]{algorithm2e}

\usepackage{makecell}

\usepackage{lipsum}

\newcommand\blfootnote[1]{%
\begingroup
\renewcommand\thefootnote{}\footnote{#1}%
\addtocounter{footnote}{-1}%
\endgroup
}

\usepackage[pagebackref=true,breaklinks=true,colorlinks,bookmarks=false]{hyperref}

\begin{document}

\title{ForgeryNet - Face Forgery Analysis Challenge 2021: Methods and Results}

\author{Yinan He, Lu Sheng, Jing Shao, Ziwei Liu, \\
Zhaofan Zou, Zhizhi Guo, Shan Jiang, Curitis Sun, Guosheng Zhang, \\ Keyao Wang, Haixiao Yue, Zhibin Hong, \\ 
Wanguo Wang, Zhenyu Li, Qi Wang, Zhenli Wang, Ronghao Xu, \\
Mingwen Zhang, Zhiheng Wang, Zhenhang Huang, Tianming Zhang, Ningning Zhao}

\maketitle

\blfootnote{$\bullet$ Yinan He and Jing Shao are with SenseTime Research and Shanghai AI Laboratory.}
\blfootnote{$\bullet$ Lu Sheng is with College of Software, Beihang University }
\blfootnote{$\bullet$ Ziwei Liu is with S-Lab, Nanyang Technological University.}
\blfootnote{$\bullet$ Zhaofan Zou, Zhizhi Guo, Shan Jiang, Curitis Sun, Guosheng Zhang, Keyao Wang, Haixiao Yue and Zhibin Hong are with Department of Computer Vision Technology (VIS), Baidu Inc.}
\blfootnote{$\bullet$ Wanguo Wang, Zhenyu Li, Qi Wang, Zhenli Wang and Ronghao Xu are with State Grid Intelligence Technology Co.,Ltd.}
\blfootnote{$\bullet$ Mingwen Zhang, Zhiheng Wang, Zhenhang Huang, Tianming Zhang and Ningning Zhao are with Safety and Security Technology Group, DiDi.}

\begin{abstract}
The rapid progress of photorealistic synthesis techniques has reached a critical point where the boundary between real and manipulated images starts to blur. Recently, a mega-scale deep face forgery dataset, \textbf{ForgeryNet} which comprised of 2.9 million images and 221,247 videos has been released. It is by far the largest publicly available in terms of data-scale, manipulations (7 image-level approaches, 8 video-level approaches), perturbations (36 independent and more mixed perturbations) and annotations (6.3 million classification labels, 2.9 million manipulated area annotations and 221,247 temporal forgery segment labels). This paper reports methods and results in the \textbf{ForgeryNet: Face Forgery Analysis Challenge 2021} which employs the ForgeryNet benchmark. The model evaluation is conducted offline on the private test set. A total of 186 participants registered for the competition, and 11 teams made valid submissions. We will analyze the top ranked solutions and present some discussion on future work directions.
\end{abstract}

\section{Introduction}
Photorealistic facial forgery technologies, especially recent deep learning driven approaches~\cite{petrov2020deepfacelab,li2019faceshifter,fried2019text}, give rise to widespread social concerns on potential malicious abuse of these techniques to eye-cheatingly forge media (\ie, images and videos, \etc) of human faces.
%
Therefore, it 
is of vital importance to develop reliable methods for face forgery analysis\footnote{In this paper, the definition of the term ``face forgery'' refers to an image or a video containing modified identity, expressions or attribute(s) with a learning-based approach, distinguished with 1) a so-called ``CheapFakes''~\cite{paris2019deepfakes} that are created with off-the-shelf softwares without learnable components and 2) ``DeepFakes'' that only refer to manipulations with swapped identities~\cite{dolhansky2020deepfake}.}, so as to distinguish \emph{whether} and \emph{where} an image or video is manipulated.

Leveraging on the ForgeryNet~\cite{forgerynet} dataset, we organize the \textit{ForgeryNet: Face Forgery Analysis Challenge 2021} (ForgeryNet Challenge) collocated with the Workshop on Sensing, Understanding and Synthesizing Humans at ICCV2021~\footnote{Workshop website: \url{https://sense-human.github.io/}.}. The goal of this challenge is to boost the research on face forensic. Specifically, the ForgeryNet is comprised of 2.9 million images and 221,247 videos, which is the largest publicly available deep face forgery dataset in terms of the numbers of the data and the manipulation methods. The dataset construction of the hidden dataset is the same as the public dataset.

In the following sections, we will describe this challenge, analyze the top ranked solutions and provide discussions to draw conclusion derived from the competition and outline future work directions.

\section{Challenge Overview}

\subsection{Platform}
The ForgeryNet Challenge is hosted on the CodaLab platform~\footnote{Challenge website: \url{https://competitions.codalab.org/competitions/33386}.}. After registering on the ForgeryNet Challenge, teams are required to upload their prediction files to the CodaLab platform for the ranking. Each team is allowed to submit their images which contain their models, and each team is allocated one 32 GB Tesla V100 GPU to perform evaluation on the hidden test set.

\subsection{Dataset}
The ForgeryNet Challenge 2021 on Face Forgery Analysis employs the ForgeryNet dataset~\cite{forgerynet} that was proposed in CVPR 2021. 
ForgeryNet is a large-scale publicly available deep face forgery dataset in terms of data-scale (2.9 million images, 221,247 videos), manipulations (7 image-level approaches, 8 video-level approaches), perturbations (36 independent and more mixed perturbations) and annotations (6.3 million classification labels, 2.9 million manipulated area annotations and 221,247 temporal forgery segment labels). 
ForgeryNet can be used to train and evaluate algorithms of face forgery analysis. 
The hidden test set is devised for the ForgeryNet Challenge, the data construction of the hidden test is as the same as the public test set. All the teams participating the ForgeryNet Challenge are restricted to train their algorithms on the publicly available ForgeryNet training dataset.

\subsection{Evaluation Metric}

\textbf{Track~1\&2.} Considering face forgery classification as binary classification, we can leverge AUC(Area Under the ROC Curve) as evaluation criteria. An ROC curve (receiver operating characteristic curve) is a graph showing the performance of a classification model at all classification thresholds. This curve plots two parameters: True Positive Rate (TPR) and False Positive Rate (FPR). Specifically, forgery class is Positive, live class is Negative.
\begin{equation}
    FPR = \frac{FP}{FP+TN}, \quad TPR = \frac{TP}{TP+TN} 
\end{equation}

Specifically, \textit{TP}, \textit{TN} and \textit{FP} correspond to True Positive, True Negative and False Negative. The AUC(Area under the ROC Curve) determines the final ranking. An ROC curve plots TPR vs. FPR at different classification thresholds. That is, AUC measures the entire two-dimensional area underneath the entire ROC curve (think integral calculus) from (0,0) to (1,1).

\textbf{Track 3.} For each video, face forensics methods to be evaluated are expected to provide temporal boundaries of forgery segments and the corresponding confidence values. We follow metrics used in ActivityNet~\cite{ghanem2018activitynet} evaluation, and employ Interpolated Average Precision (AP) as well as Average Recall@2(AR@2) for evaluating predicted segments with respect to the groundtruth ones. To determine if a detection is a true positive, we inspect the temporal intersection over union (tIoU) with a ground truth segment, and check whether or not it is greater or equal to a given threshold(e.g. tIoU$>$0.5).

\subsection{Timeline}
The ForgeryNet Challenge lasted for ten weeks from July 12, 2021 to September 19, 2021. During the challenge, participants had access to the public ForgeryNet dataset, and they are restricted to used the public ForgeryNet training dataset for the training of their model. The challenge results were announced in October 17, 2021. A total of 186 participants registered for the competition, and 11 teams made valid submissions.

\section{Results and Solutions}
Among the 11 teams who made valid submissions, many participants achieve promising results. We show the final results of the teams in Table~\ref{table:all_result}. In the following sections, we will present the solutions of top-3 teams.

\setlength{\tabcolsep}{5pt}
\begin{table}[t]
\centering
\ra{1.1}
\caption{Final results of the top teams in the ForgeryNet Challenge 2021.}
\vspace{3pt}
\label{table:all_result}
\begin{tabular}{c|cccc}
\toprule
Track & Ranking & Team name  & Score                               & Time  \\ \hline
1     & 1       & asdfqwer   & 0.969187578                         & 32min \\
2     & 1       & asdfqwer   & 0.949600966                         & 37min \\ 
3     & 1       & asdfqwer   & 0.569632581                         & 31min \\
3     & 2       & sgai\_lab  & 0.526588228                         & 33min \\
3     & 3       & DiDi\_SSTG & 0.069723898                         & 84min \\ \bottomrule
\end{tabular}
\end{table}

\subsection{Solution of First Place}
\textit{Team members: Zhaofan Zou, Zhizhi Guo, Shan Jiang, Curitis Sun, Guosheng Zhang, Keyao Wang, Haixiao Yue, Zhibin Hong}
\begin{figure}[h!]
\centering
\includegraphics[width=0.45\textwidth]{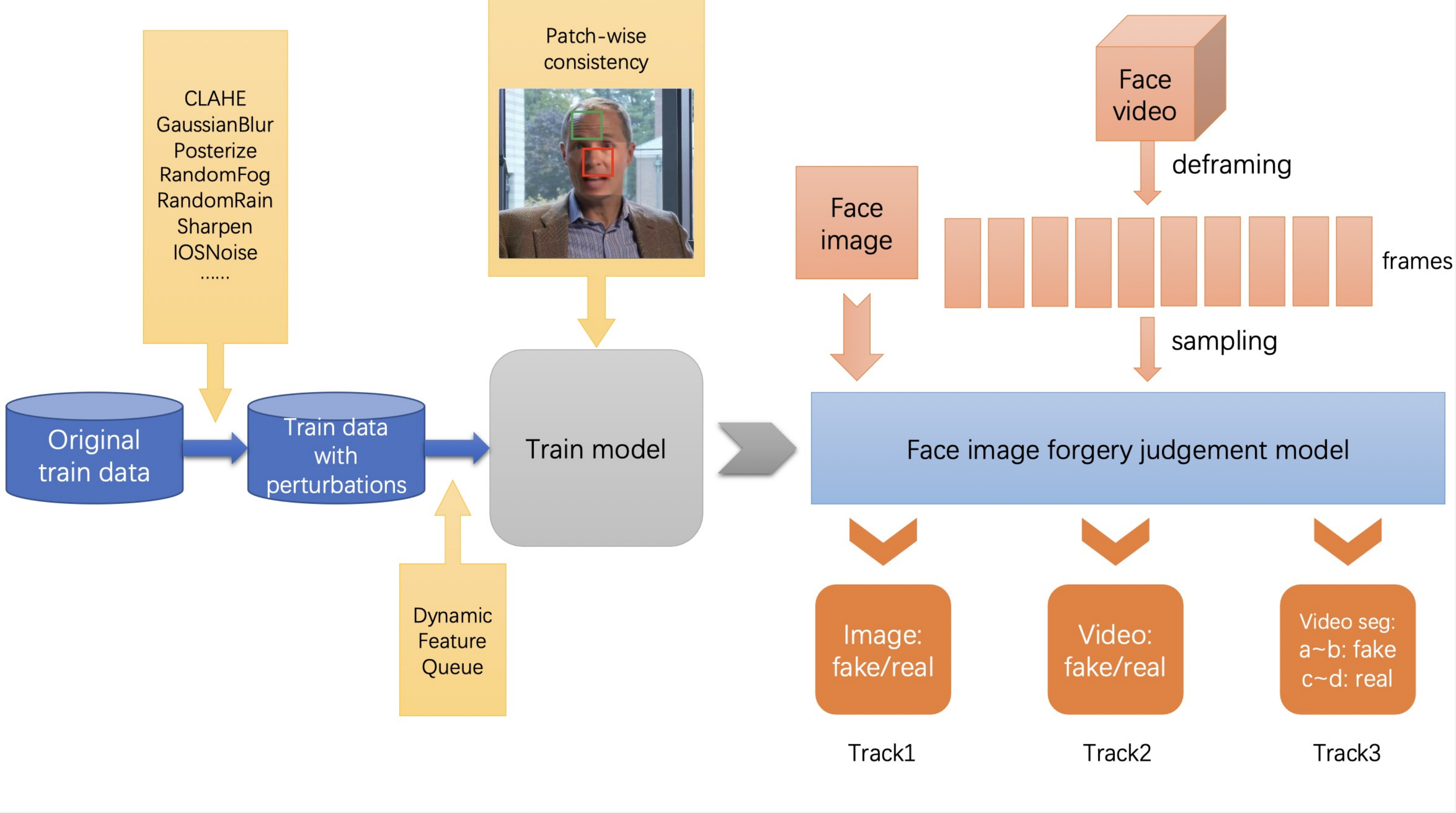}
\caption{The pipeline of the first-place solution.}
\label{fig:1_1}
\end{figure}

\begin{figure}[h!]
\centering
\includegraphics[width=0.45\textwidth]{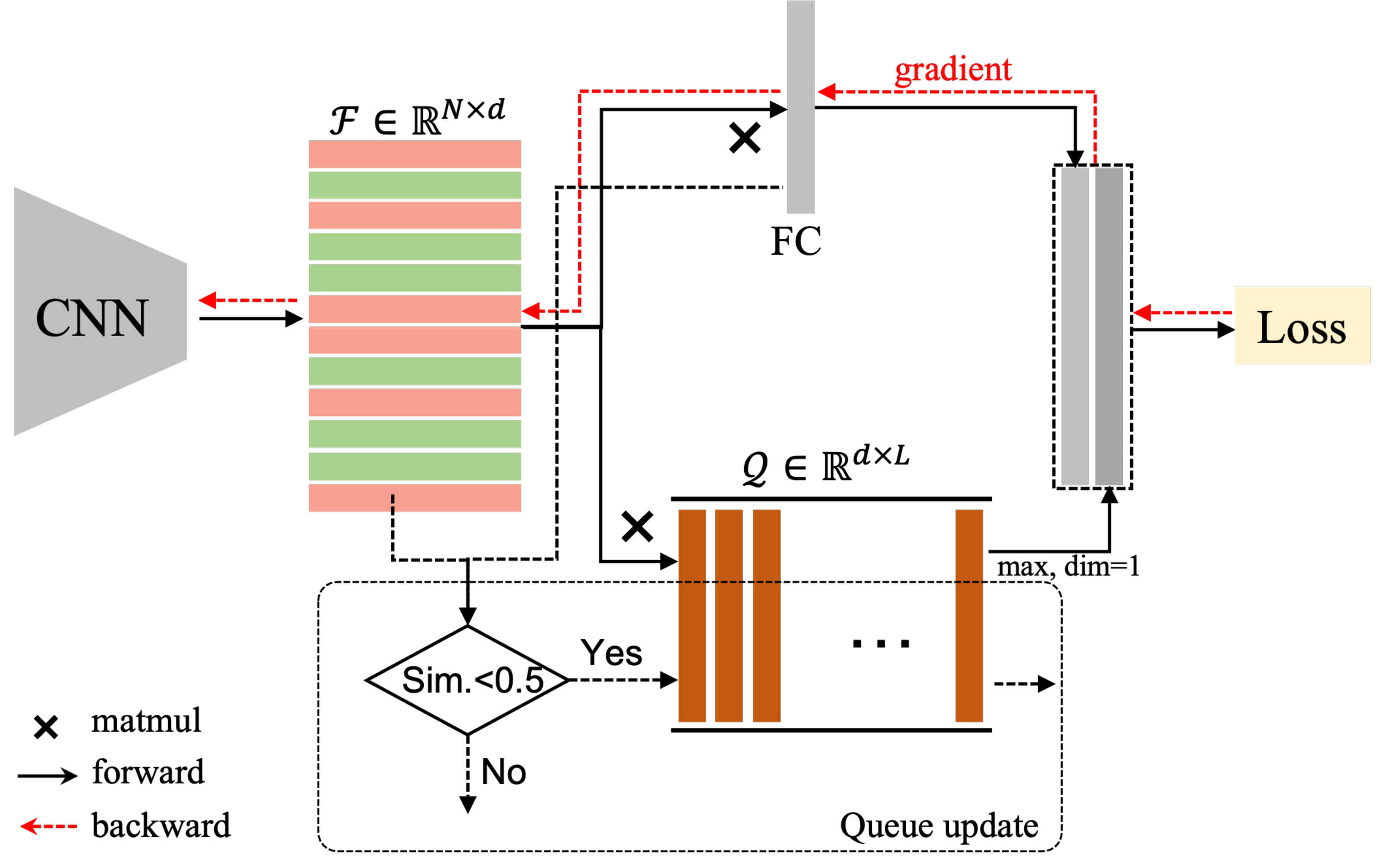}
\caption{The diagram of dynamic feature queue.}
\label{fig:1_2}
\end{figure}

\noindent \textbf{General Method Description.}
The champion team propose a robust method for face forgery analysis. It has two components:
\textbf{1) dynamic feature queue (DFQ)}: which provide a more global metric space for metric learning and ensure the stability of model training.
\textbf{2) global patchwise consistency}: which is adopted as a cue to identify forgery images/videos without face detection from different level features.
\textbf{3) process for video track}: they achieve high-speed forgery verification and temporal localization of the video by discretely sampling the video.

\noindent \textbf{Training Description:} Considering the trade-off between performance and efficiency, they use ResNeSt-50 as the backbone of the 16-way classification for three tracks. To shorten the time-consuming of the overall pipeline, they discard the step of face detection by directly analyzing the whole image/frame, and Fig.~\ref{fig:1_1} is their whole pipeline. First, they apply the face Pair-Wise Consistency~\cite{zhao2021learning} to the analysis of the entire image, not just the face area. Different features from the backbone’s multi-stage are computed the pair-wise similarity scores of all possible pairs of local patches. The similarity scores would be used to do a 16-way classification. Second, the Pair-Wise Consistency model would be regarded as a new pretrained model. As shown in Fig.~\ref{fig:1_2}, they set up a dynamic queue for the forgery feature. By calculating the cosine similarity between the feature in the batch and the queue feature, the pair with the closest feature distance is brought closer to form multiple clusters. A sufficiently long queue can help them solve the Catastrophic Forgetting and fully mining hard samples. Also, they set a learnable positive sample center feature and draw all positive samples closer to this feature, which will make the positive sample features sufficiently clustered in the feature space.

\begin{figure}[h!]
\centering
\includegraphics[width=0.45\textwidth]{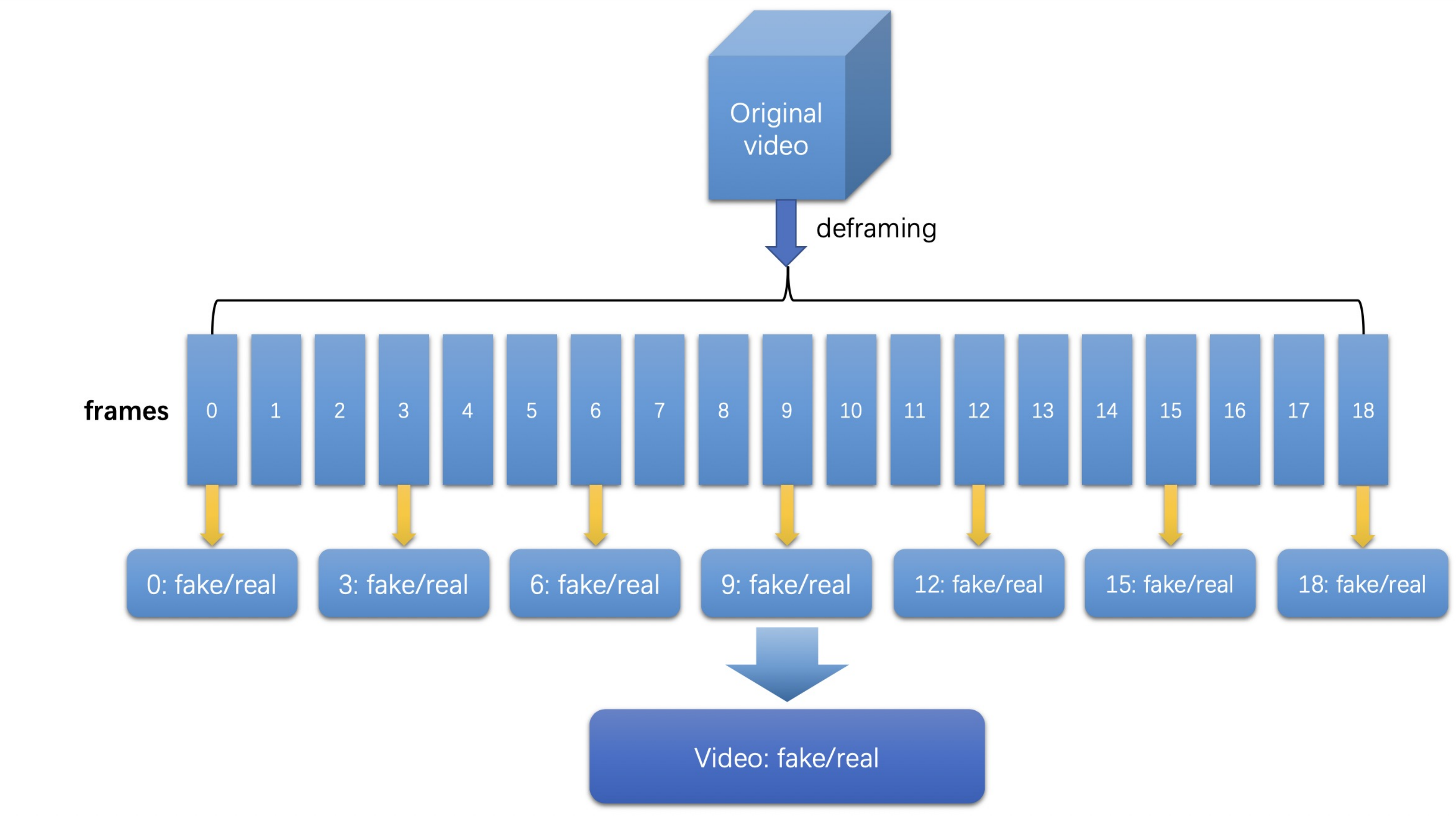}
\caption{Track2 Post-processing of the first-place solution.}
\label{fig:1_3}
\end{figure}

\begin{figure}[h!]
\centering
\includegraphics[width=0.45\textwidth]{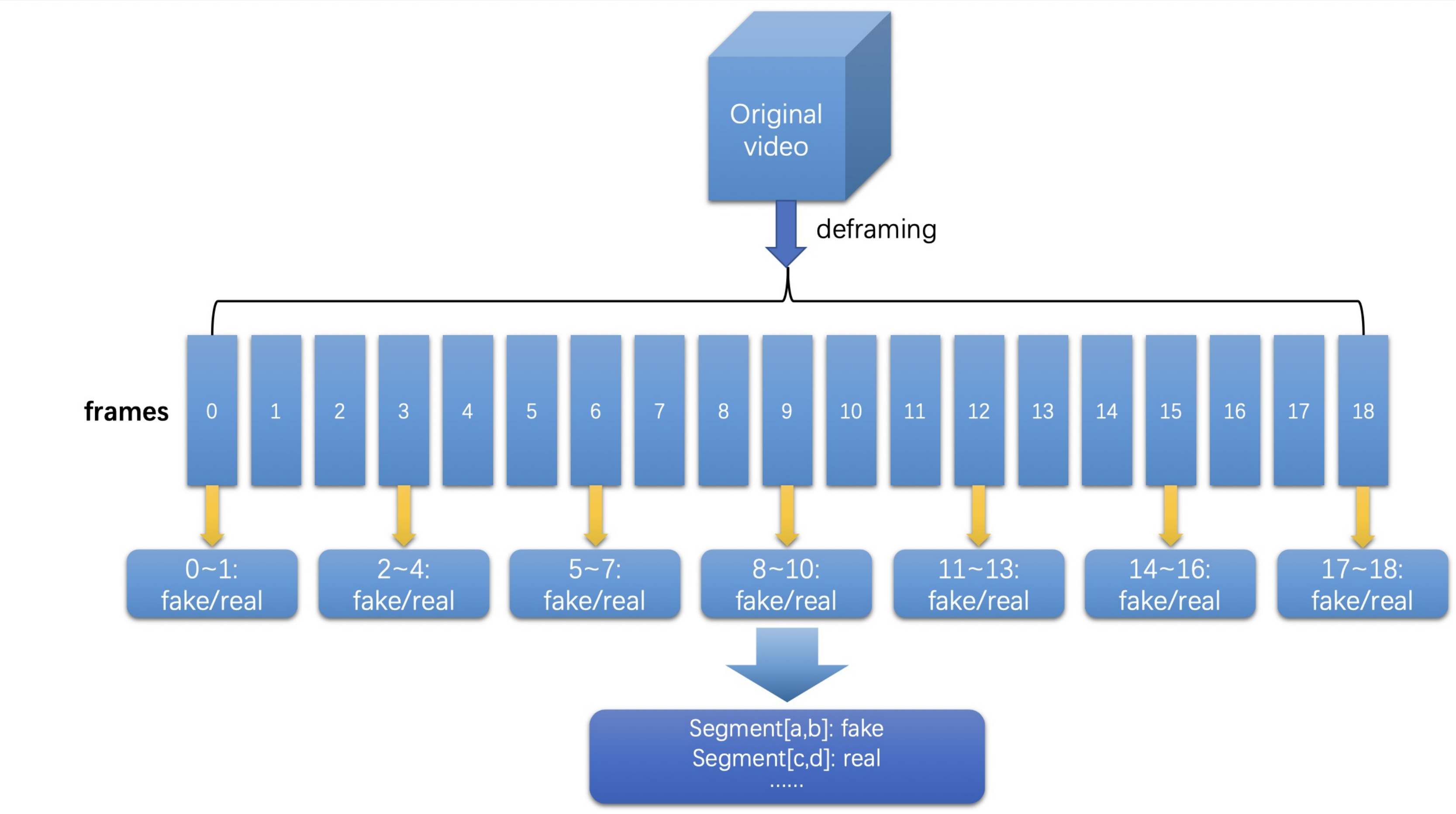}
\caption{Track3 Post-processing of the first-place solution.}
\label{fig:1_4}
\end{figure}

\noindent \textbf{Testing Description:} For Track1, they only use DFQ and patch-wise consistency as auxiliary branches and do not use them in testing. Only the feature of the backbone is selected for the 16-way classification.
Fig.~\ref{fig:1_3} and Fig.~\ref{fig:1_4} show that videos are disassembled into frames with the OpenCV library for Track2 and Track3. They uniformly sample 16 frames of each video in Track2 and Track3 so that while balancing time-consuming and precision, at least one fake frame can be sampled with the greatest probability. They could get the fake probability of each frame by classification model inference. They set a threshold 0.5 to judge the real or fake of the frame and use the average score of all fake frames as the video fraud probability. If there are no fake frames in the video, the average score of all frames would be regarded as video fraud probability. For forgery video temporal localization, if the n-th frame is fake and the stride of the video is k, the fake segment is [n $-$ stride/2, n $+$ stride/2].

\noindent \textbf{Implementation Details:} They normalize all images/frames with mean [0.5, 0.5, 0.5] and standard deviation [0.5, 0.5, 0.5], and resize them to 224$\times$224. They also use more data augmentations\cite{buslaev2020albumentations} over 40, including JPEG-compression, Gaussian noise/blur, brightness contrast, random erasing, and color jittering, to mimic distortions caused by compression and packet loss during transmission to improve the generalization of developed models. For data augmentation, the with 99\% probability randomly select one perturbation from some set of perturbation methods and apply it to the input image. They adopt synchronized Adam training on 2 A100-GPUs. A minibatch involves 128 images per GPU, and the ratio of real to fake in each mini-batch is 1:1. They use a weight decay of 0.0005 and betas of [0.9, 0.999]. The learning rate is 0.0002 for the first three epochs, and they adopt a multi-step learning rate schedule with 15 epochs, and the learning rate is decreased by a factor of 0.7 at every epoch.


\subsection{Solution of Second Place}
\textit{Team members: Wanguo Wang, Zhenyu Li, Qi Wang, Zhenli Wang, Ronghao Xu}

The global feature of image is used to classify directly where the face detection step is omitted. They model can provide 2-3x speedup with the similar precision. 

\noindent \textbf{Training Description.}
They adopt ResNeSt-50 initialized with pretrained weights on ImageNet~\cite{deng2009imagenet} as backbone and train a frame-based 16-way classifier to accomplish the temporal localization task (track 3). Additionally, an asymmetric regression branch is involved to assist the model in learning pixel-level semantic features.

\begin{figure}[t]
\centering
\includegraphics[width=0.45\textwidth]{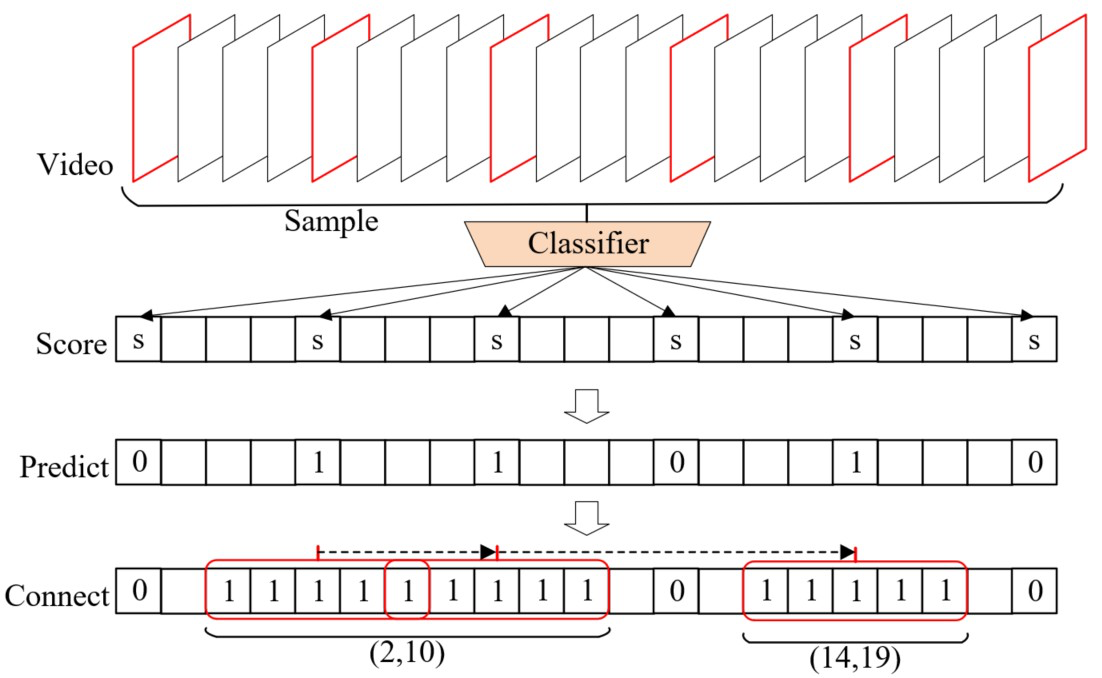}
\caption{The diagram of the second-place solution.}
\label{fig:2_1}
\end{figure}


\noindent \textbf{Testing Description.}
During testing, they uniformly sample 1 frame every 10 frames from each video and make binary classiﬁcation (real or fake) for each frame to get scores of the fake sample. Finally, as shown in Figure 1, they get the result of the temporal localization by using the sliding window strategy.

\noindent \textbf{Implementation details.}
They normalize the images with mean 0 and standard devation 1 and resize them to 224 $\times$ 224. To further enhance the generalization ability of our method, they carefully select several types of data augmentation to mimic distortions in private test set. The model is first trained for 50 epochs using Adam~\cite{adam} optimizer with batch size 128, betas 0.9 and 0.999, and weight decay of $5 \times {10}^{-4}$. The learning rate is initialized to $1 \times 10^{-3}$ and cut by 0.6 every 10 epochs. After that, they reload the best model from first step training and retrain the model for 50 epochs with richer data augmentation (\textit{SomeOf}$ (0,5) $ turn to \textit{SomeOf}$ (0,8) $). Model spends 300 hours for training with 8 GPUs (each with 32 GB memory consumption).

\subsection{Solution of Third Place}
 \textit{Team members: Mingwen Zhang, Zhiheng Wang, Zhenhang Huang, Tianming Zhang, Ningning Zhao}
 
\begin{figure}[h]
\centering
\includegraphics[width=0.2\textwidth]{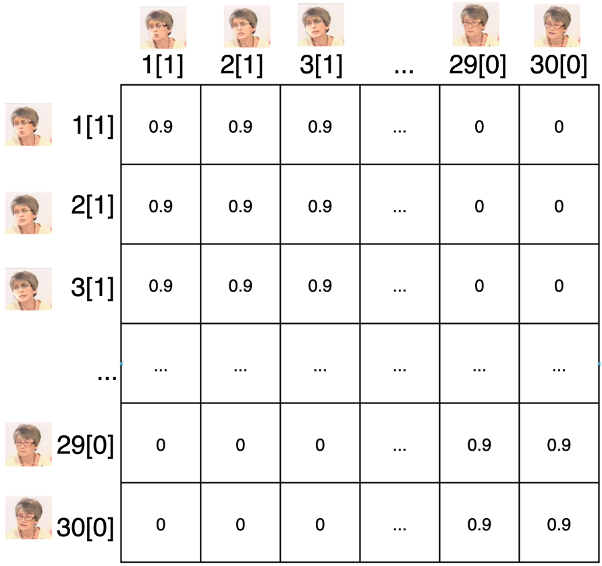}
\caption{Similarity matrix.}
\label{fig:3_1}
\end{figure}

\begin{figure}[h!]
\centering
\includegraphics[width=0.45\textwidth]{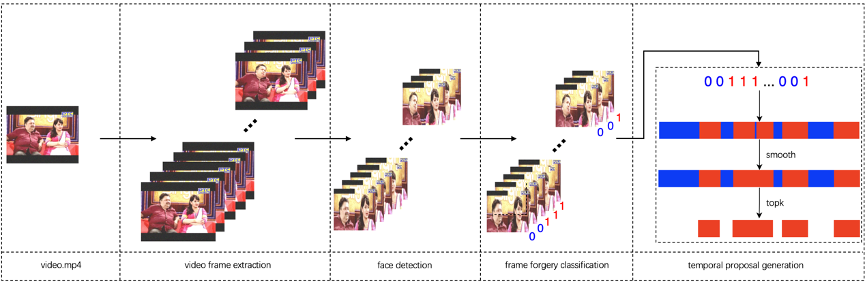}
\caption{The pipeline of the third-place solution.}
\label{fig:3_2}
\end{figure}

Recently, transformer-based methods have made good achievements in various fields of computer vision. 
Self-attention mechanism has strong long-distance modeling capability and can effectively overcome the limitations of convolution. 
However, there are also some shortcomings, such as slower convergence speed and more training data than CNNs. 
VOLO introduces a new outlook attention and presents a simple and general architecture, which can efficiently encode the fine level features and context information into token representation without using extra training data. 
They make some improvements based on VOLO, and the new method is suitable for video scenes. 
VOLO introduces a simple and lightweight attention mechanism called outlook, this new attention mechanism can enrich the fine-grained feature information in token and enhance the spatial feature expression ability of the model. On the basis of VOLO, we introduce self-attention in the temporal dimension, which can build a long-distance similarity relationship between different video frames. 

For instance, as shown in Fig.~\ref{fig:3_1}, 30 frames are extracted from the video, real and fake frames are labeled as 0 and 1 respectively. They calculate the similarity between all video frames, and video frames with the same label have a high similarity which values 0.9, while different kinds of video frames have a low similarity which values 0.

the pipeline of video VOLO for ForgeryNet is summarized in Fig.~\ref{fig:3_2}. In order to improve the speed as much as possible on the premise of good performance, they extract key frames from full video and make face detection, only a small face region is used as the input of forgery classification model. After model recognizes the class of each frame, they use manual rules to generate temporal proposals.

\noindent \textbf{Data augmentation:} In order to improve the generalization of the model, a variety of data augmentation methods are used, including GaussianBlur, MedianBlur, GaussNoise, ISONoise, RandomBrightness, ImageCompression, HueSaturationValue, InvertImg, RandomGamma, RandomContrast, CLAHE, Cutout, etc.

\noindent \textbf{Video frame extraction:} For the task of temporal forgery localization, video frames in the same segment have high similarity, which is a key feature for model learning, while the action feature is relatively less important.

In order to ensure the model can learn similarity features effectively, they sample 30 frames from the video as the input of model according to a certain rule. For real video, 30 consecutive frames are randomly sampled, and those less than 30 frames are filled with the last frame. For forgery video, the proportion of fake frame is $\alpha(0< \alpha < 1)$, and real frame is $1 - \alpha$, insufficient real or fake frames are filled with the last frame.

\noindent \textbf{Face detection} As manipulated areas of fake frames concentrate on faces, they will detect faces with Retinaface an effective face detection method before video frames are input to model to reduce background interference. Most faces in frames can be detected effectively, a few can’t be detected will be estimated with the front and back frames. If there are faces still can’t be obtained, the full image is used to replace the face box. For the completeness of face region, face box is randomly expanded outward between 0.2h and 0.6h, where h is the height of the original face box.

If there are four or more faces in a video frame, only the four boxes with the largest area will be retained, and the remaining faces will be ignored. If the number of faces is less than 4, all faces will be retained. The remaining faces are merged into an image with resolution 160×160.

\noindent \textbf{Frame Forgery classification.}
They make some improvements based on VOLO, attention mechanism in the temporal dimension is introduced and the new method is suitable for video scenes. In image recognition task, transformer-based methods divide the image into multiple patches, and different patches learn the similarity in spatial dimension through attention mechanism. In the task of temporal forgery localization, not only the similarity of spatial dimension between different patches in the same frame is important, but also the similarity of temporal dimension between different frames is also necessary. Therefore, they extend the modeling capability for temporal similarity based on VOLO, and the improved method can be applied to the scene of video understanding.

Video frames with the same label have high similarity, while those with different label have low similarity. Each frame is divided into multiple disjoint patches, and each patch is encoded to a token, then they can get a similarity matrix calculated with the token at the same position in different video frames. The similarity matrix means self-attention in temporal dimension, which is useful for video understanding. Besides, they also implement self-attention in spatial dimension with different tokens in the same frame. They can classify video frames with effective features including both temporal and spatial information.

For the training data, the similarity of video frames with the same label values 0.9, and the similarity of those different video frames values 0. In this way, a similarity matrix with resolution 30×30 will be generated, and it is used as the label to calculate cross entropy loss. Another cross entropy loss is calculated with the recognition result of each video frame, and the total loss is the weighted summation of these two losses.

\noindent \textbf{Testing description.} In testing stage, 30 frames are extracted from the whole video at equal intervals, and those less than 30 frames are filled with the last frame. They only sample 10 frames from 30 frames at equal intervals to detect faces. If there are faces still can’t be obtained, the full image is used to replace the face box. For the completeness of face region, face box is randomly expanded outward 0.4h, where h is the height of the original face box.
 
30 video frames with faces are input to model, and they will get the class label of each frame. To get all video frame labels, those frames not recognized by the model are assigned to the label of the nearest frame in front of them.

\noindent \textit{Temporal proposal generation.} The model predicts fake score of each frame, if the score is greater than the threshold, it is considered as a fake frame, otherwise it is a real frame. After each frame is labeled, video frames with the same label are merged into the same segment. If the length of real segment is less than 4, all frames in the segment will be labeled as fake frame. Among all segments with a length greater than 10, they select the top 4 as temporal proposals. In our method, three sets of temporal proposals will be generated based on three thresholds, and we get the final results by doing NMS with all temporal proposals.

\section{Discussion}
The winning solutions mentioned above have achieved promising results on the ForgeryNet Challenge, these solutions focus on different aspects in developing a robust and efficient face forensic model. To briefly sum up, among their solutions, there are two key points are essential for improving the performance of the face forensic task. 
\textit{1) Forensic Model:} Besides the commonly used deep learning models, such as ResNeSt, these solutions not only inherit the models which are published recently, but also devise novel framework for detecting forgery cues, such as global patchwiseconsistency in the first place solution. 
\textit{2) Post-processing:} These winning methods leverage on different post-processing strategies to boost their model performance while balancing the time-consuming. 
Moreover, we believe that there is still much room for improvement in the future face forensic challenge.  For example,  \textit{1) Size:} The size of public test set could be larger in the future.  \textit{2) Diversity:} The forgery images could be more realistic.
\vspace{8pt}

\noindent \textbf{Acknowledgments.}We sincerely thank the codebase from CelebA-Spoof Challenge~\footnote{\url{https://competitions.codalab.org/competitions/26210}.}, especially for the helpful discussions from Yuanhan Zhang.


{\small
\bibliographystyle{ieee_fullname}
\bibliography{challenge/sections/bibliography}
}

\end{document}